\documentclass[letterpaper, 10 pt, conference]{ieeeconf}

\IEEEoverridecommandlockouts
\usepackage{graphics} 
\usepackage{epsfig} 
\usepackage{times} 
\usepackage{amsmath} 
\usepackage{amssymb}  
\usepackage[table]{xcolor}
\usepackage[ruled,linesnumbered, noend]{algorithm2e}
\usepackage{flushend} 

\usepackage{soul}
\usepackage{cite}
\usepackage{comment}
\usepackage{multirow}
\usepackage{graphicx}
\usepackage{wrapfig}
\usepackage{soul, color}
\usepackage{wrapfig}
\usepackage{mathtools}
\usepackage{graphicx}
\usepackage{booktabs}
\usepackage{adjustbox}

\usepackage{caption}
\usepackage{subfigure}
\usepackage{tabularx}  
\usepackage{booktabs}  

\definecolor{LightCyan}{rgb}{0.88,1,1}
\definecolor{Gray}{gray}{0.9}

\newcommand{\ve}[1]{\mathbf{#1}} 
\newcommand{\tve}[1]{\tilde{\mathbf{#1}}} 

\title{Kinesthetic-based In-Hand Object Recognition\\ with an Underactuated Robotic Hand}

\author{Julius Arolovitch$^*$, Osher Azulay$^*$ and Avishai Sintov
\thanks{$^*$ These authors contributed equally.}
\thanks{J. Arolovitch is with the Robotics Institute, Carnegie Mellon University, Pittsburgh, PA, USA. E-mail: jarolovi@andrew.cmu.edu}
\thanks{O. Azulay and A. Sintov are with the School of Mechanical Engineering, Tel-Aviv University, Israel. E-mail: osherazulay@mail.tau.ac.il, sintov1@tauex.tau.ac.il}
}

\begin{document}

\setlength{\belowdisplayskip}{2pt}
\setlength{\belowdisplayshortskip}{3pt}
\setlength{\abovedisplayskip}{2pt} 
\setlength{\abovedisplayshortskip}{3pt}
\setlength{\parskip}{0pt}


\maketitle
\thispagestyle{empty}
\pagestyle{empty}


\begin{abstract}
    Tendon-based underactuated hands are intended to be simple, compliant and affordable. Often, they are 3D printed and do not include tactile sensors. Hence, performing in-hand object recognition with direct touch sensing is not feasible. Adding tactile sensors can complicate the hardware and introduce extra costs to the robotic hand. Also, the common approach of visual perception may not be available due to occlusions. In this paper, we explore whether kinesthetic haptics can provide in-direct information regarding the geometry of a grasped object during in-hand manipulation with an underactuated hand. By solely sensing actuator positions and torques over a period of time during motion, we show that a classifier can recognize an object from a set of trained ones with a high success rate of almost 95\%. In addition, the implementation of a real-time majority vote during manipulation further improves recognition. Additionally, a trained classifier is also shown to be successful in distinguishing between shape categories rather than just specific objects.
    

\end{abstract}

\section{Introduction}

Underactuated hands, characterized by their ability to conform to object shapes through compliance, offer simplicity and cost-effectiveness \cite{Odhner2011,Azulay2023}. This is in contrast to traditional robotic hands, while precise and proficient, are limited by their intricate design, high costs and complex control schemes \cite{Bai2014,Michalec2010}. Underactuated hands have been shown to provide stable grasps through open-loop control \cite{Dollar2010} and an ability to perform precise in-hand manipulation tasks \cite{Calli2016, Calli2018a,Sintov2019}. Some attempts have been made to also enable object recognition with underactuated hands \cite{Gandarias2018,Zhou2022}. These, however, often require additional sensory hardware on the hand \cite{liarokapis2015unplanned,Jiao2020,da2022tactile}.

\begin{figure}
    \centering
    \includegraphics[scale=.2]{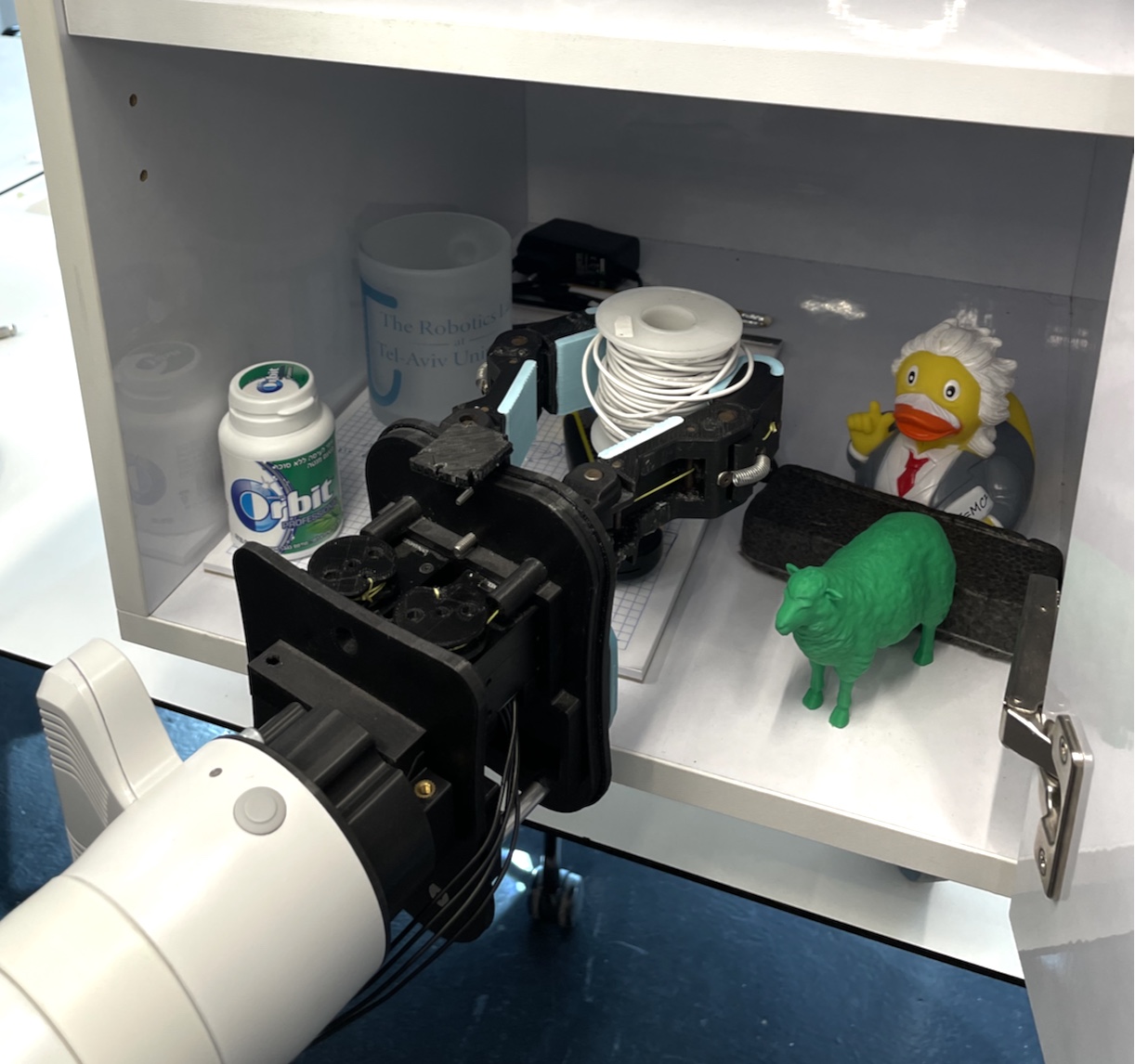}
    \caption{\small In-hand recognition of an object in an occluded environment with an underactuated hand where visual perception is not available. The hand relies on kinesthetic perception during in-hand manipulation to either recognize the specific object from a trained set or its general shape.}
    \label{fig:intro}
\end{figure}

The recognition of an object grasped within a robotic gripper or hand has been a topic for numerous works \cite{Navarro2012,Watanabe2017}. An ability to recognize object without visual perception increases the workspace of the robot and enables it to work in regions without proper lighting or line-of-sight \cite{azulay2022learning}. Relying solely on continuous visual feedback limits the performance in various tasks where visual uncertainty or occlusion may occur. In such cases, it may be impossible to solve the task altogether. Examples include grasping an arbitrary object at the back of a cabinet or in confined spaces as demonstrated in Figure \ref{fig:intro}.

The common approach for object recognition is the use of haptics or, in particular, tactile sensing \cite{Rouhafzay2020,Pastor2021}. In such sensing, tactile sensors have direct contact with the surface of the object and provide a glance of its geometry for recognition inference \cite{Drimus2014}. Often, the hand requires multiple tactile glances in order to provide a certain recognition regarding the object \cite{Jin2013,Kirby2022}. However, adding tactile sensors to robotic hands can be expensive and may complicate the hardware. Kinesthetic haptics, on the other hand, gather information by sensing movement, force and position of actuator joints in the hand \cite{Carter2005}. Prior work  has combined tactile images along with kinesthetic information while utilizing $k$-means clustering to increase classification success rate \cite{Luo2018}. More recently, kinesthetic haptics was used to recognize objects by observing finger kinematics during multiple grasps \cite{sintov2023simple}. However, these assume rigid hands and full knowledge of the kinematic configuration of the fingers.

Object recognition using underactuated hands has mostly relied on either visual methods or tactile perception \cite{li2020skin,flintoff2018single,Huang2022,Hanson2022,Cao2022}. In \cite{spiers2016single}, embedded force sensors along the two fingers of an underactuated hand were introduced for object classification and feature extraction through single-grasp interactions. Similarly, optical tactile sensors, which observe deformation on the contact pad through an internal camera, were used on a three-finger underactuated hand for recognition of grasped objects \cite{james2021tactile}. Integrating these sensors into compliant hands adds extra costs and computational complexity to the design. This is even more crucial for open-source 3D printed underactuated robotic hands that attempt to provide a simple, low-cost and accessible solution \cite{Ma2017YaleOP}.  

In this paper, we address the problem of proprioceptive object recognition with underactuated hands by exploring it solely through kinesthetic haptics. Proprioception, or kinaesthesia,  offers a unique way to recognize the state of the hand-object system without relying on conventional visual or tactile cues. We focus on a tendon-based underactuated hand where actuators at the base of the hand pull tendons running along the fingers, which also have passive joints and springs. With no tactile sensors on the fingers and no exact model for these compliant mechanisms \cite{Sintov2019}, we observe signals from the actuators during in-hand manipulation of various objects and explore whether they embed crucial information regarding their shapes. The feature state representation and data-based modeling are explored.

A single proprioceptive state signal from the hand at some time instant is not sufficient for representing the shape of a grasped object. Therefore, we hypothesize that a sequence of states during in-hand manipulation could embed geometric information such as curvature, flat surface and corners. Consequently, we collect sequential data during the manipulation of various objects. With the data, our investigation delves into an extensive analysis of various classification models, either simple or temporal-based, that capture the intricate relationships between the hand's proprioceptive data and the underlying object geometries. Furthermore, we utilize a simple majority vote paradigm for increasing the certainty of the recognition in real-time during manipulation. Finally, the generalization capabilities of the models are explored, evaluating their performance on generalizing to categories of shapes rather than specific objects in the training set.

By focusing on an object recognition model without vision or tactile sensing, we provide a novel capability that aligns with simplicity and cost-effectiveness in hardware design. Without additional sensors, a low-cost underactuated robotic hand can pick-up an object and simply manipulate it between the fingers for a short period of time in order to recognize it. Hence, by circumventing the limitations of visual and tactile perception, we offer a pure algorithmic and potentially more robust approach to in-hand object recognition. While out of the scope of this work, the detection for a successful grasp at the pick-up phase can also be recognized through kinesthetic haptics. We note also that the proposed approach can be complementary to visual perception in order to increase recognition certainty or replace it entirely if a line-of-sight is not available.

\section{Method}
\subsection{Problem Definition and Approach}

\begin{figure}
    \centering
    \includegraphics[scale=.275]{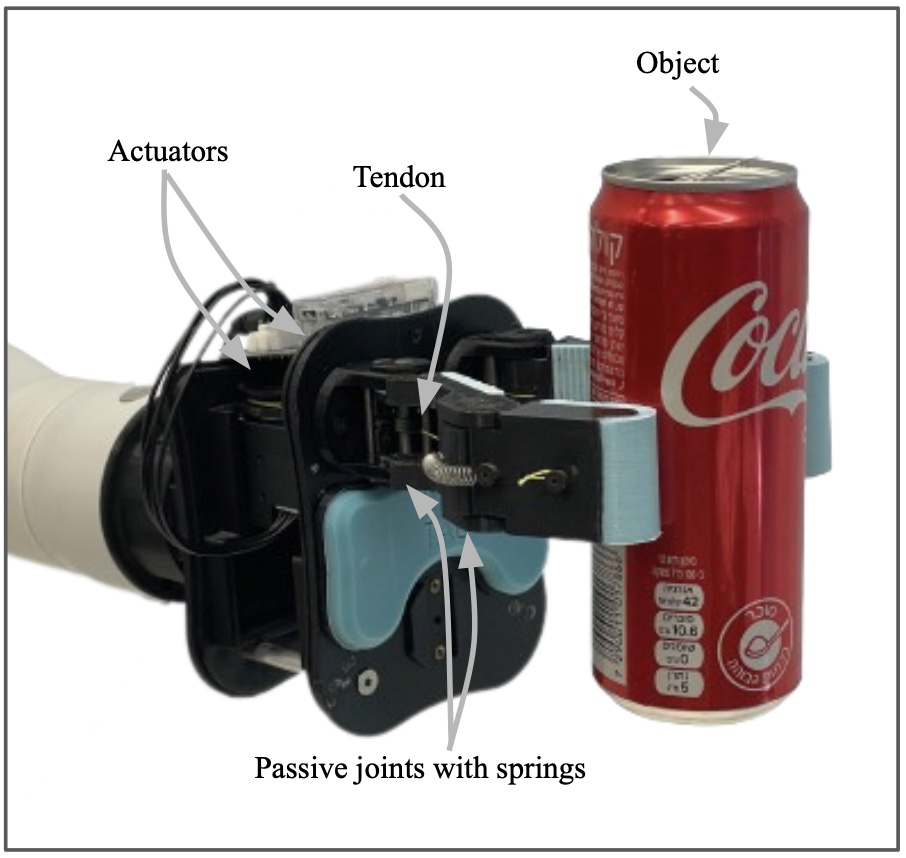}
    \caption{\small An underactuated robotic hand (OpenHand Model-O \cite{Ma2017YaleOP}) with two-fingers. The hand is mostly 3D printed and is tendon-based. Each finger consists of two passive joints with springs. The tendons run along the length of the two fingers and are pulled by two actuators at the base of the hand. }
    \label{fig:hand}
\end{figure}

Consider a two-finger underactuated hand as seen in Figure \ref{fig:hand}. The hand comprises of two opposing tendon-based fingers such that a manipulated object between the fingers performs a planar motion \cite{Sintov2019}. Each finger consists of two compliant joints with springs, and a tendon runs along its length. The tendon is connected to an actuator situated at the base of the hand and the finger flexes upon pulling of the tendon. The distal links of the fingers are equipped with high friction pads to prevent slipping. The hand has no tactile sensors on the fingers and no visual perception of the hand-object system is available. The hand can only measure kinesthetic features such as actuator angles and torques.

Let $\ve{x}\in\mathcal{C}$ represent the observable state of the hand where $\mathcal{C}\subset\mathcal{R}^n$ is some $n$-dimensional state space. Similarly, vector $\ve{a}\in\mathcal{U}$ denotes an action exerted on the hand where $\mathcal{U}\subset\mathbb{R}^{2}$ is the action space. The action vector $\ve{a}$ corresponds to angle changes of the two actuators, signifying tendon pull or release, over a fixed time step $\Delta t$. Since no visual perception is used, the true state of the hand and the pose of the object $\ve{T}\in SE(2)$ with respect to the hand are not known and cannot be observed directly \cite{azulay2022learning}.

Given a set of $m$ objects $\{\mathcal{O}_1,\ldots\mathcal{O}_m\}$, it is required to identify an object grasped by the hand from the set without any use of visual feedback nor tactile sensors. The query object will be identified during the grasp without releasing and regrasping it. It is assumed that all grasped objects are rigid. Furthermore, there is no assumption on the pose $\ve{T}_0$ of the object during the initial grasp. The main objective of this work is, therefore, twofold. First, we aim to identify the set of $n$ kinesthetic features that best represent relevant information regarding the shape of the grasped object. Second, we wish to utilize the features and explore the ability of a data-based model to classify grasped objects. Consider state measurements arriving sequentially $\{\ve{x}_1,\ldots,\ve{x}_k\}$ in real-time while manipulating a given query object $\mathcal{O}$ whose class is unknown. It is required to solve the following maximization problem \\
\begin{equation}
    \label{eq:P(O)}
    j^*=arg\max_j P(\mathcal{O}_j|\ve{x}_1,\ldots,\ve{x}_k)
\end{equation}
where $P(\mathcal{O}_j|\ve{x}_1,\ldots,\ve{x}_k)$ is the conditional probability for $\mathcal{O}$ to be in class $\mathcal{O}_j$.


\subsection{Data Representation}

Data is collected by recording various state transitions during in-hand manipulation of the $m$ objects. An observable state of the hand during manipulation is a kinesthetic measurement of internal features. In the tested hand, the state can consist of actuator torques and angles. In addition, we consider the case where the applied action at the given time is included in the state. Hence, the state $\ve{x}$ can be of dimension up to $n=6$. The collection of data is conducted in an episodic manner where, in each episode, an object is manipulated by exerting a set of pre-defined actions until drop. Hence, data for episode $i$ is a set of states $\mathcal{E}_i=\{\ve{x}_0,\ldots,\ve{x}_z\}$ and the corresponding object label $y_i\in\{1,\ldots,m\}$. In order to overcome biases throughout different episodic sessions, the initial state of an episode is subtracted from all states in the episode, i.e., $\tve{x}_j=\ve{x}_j-\ve{x}_0$. Subsequently, an optimized low pass filter is applied to each episode to alleviate sensor noise. Consequently, dataset $\mathcal{D}$ is of the form
\begin{equation}
    \mathcal{D}=\{(\tilde{\mathcal{E}}_i,y_i)\}_{i=1}^{N_e}
\end{equation}
consisting of $N_e$ episodes and where $\tilde{\mathcal{E}}_i=\{\boldsymbol{0},\tve{x}_1,\ldots,\tve{x}_z\}$.


\subsection{Data-based Model and Inference}

The recognition of a grasped object without tactile sensors cannot be done with only a single kinesthetic state $\ve{x}_i$. Such state will not have sufficient information regarding the shape of the object. Alternatively, state change during manipulation may have embedded geometric information such as corners and surface curvature across the object. Hence, we hypothesize that a sequential set of states $\{\ve{x}_1,\ldots,\ve{x}_k\}$ would enable the training of a model for the desired recognition. 

Sequential data enables to learn the motion pattern of the hand-object system over some period of time. Given a set of $w$ sequential states along a motion time frame $S_w=\{\ve{x}_1,\ldots,\ve{x}_w\}$, we search for a data-based classification model that could provide a solution for \eqref{eq:P(O)}. In practice, we aim to train a model $\Gamma_\psi:\mathcal{C}\times\ldots\times\mathcal{C}\to[0,1]^m$, where $\psi$ is the vector of trained parameters, to provide a probability distribution over the $m$ possible objects in the form
\begin{equation}
    \label{eq:Gamma}
    \Gamma_\psi(S_w)=\left[ P(\mathcal{O}_1|S_w),\ldots, P(\mathcal{O}_m|S_w) \right].
\end{equation}
Then, the predicted object would be $\mathcal{O}_j$ with 
\begin{equation}
    \label{eq:Gamma_max}
    j^*=arg\max_j \Gamma_\psi(S_w)]
\end{equation}
where $arg\max_i\ve{v}$ returns the index of the component in vector $\ve{v}$ with the maximum value. In order to train model $\Gamma_\psi$, dataset $\mathcal{D}$ is modified to include labeled motion sequences of pre-defined length $w$. Along each episode $\tilde{\mathcal{E}}_i$ of length $z$, a window of length $w$ is moved to generate $z-w+1$ sequences $\{S_{w,i}^{(1)},\ldots,S_{w,i}^{(z-w+1)}\}$. Each sequence $S_{w,i}^{(j)}$ is labeled with the original label $y_i$ yielding a training set $\mathcal{P}=\{(S_{w,i},y_i)\}_{i=1}^N$.

Training a sequential model with $\mathcal{P}$ for $\Gamma_\psi$ in \eqref{eq:Gamma} can be done with data-based architectures designated for sequence or temporal modeling, such as the Long Short-Term Memory (LSTM) \cite{Yu2019} and Temporal Convolutional Network (TCN) \cite{Bai2018}. LSTM is a class of Recurrent Neural-Networks (RNN) aimed to learn sequential data and is able to selectively retain or discard information from previous time steps making it well-fit for long-term dependencies. TCN is a type of neural-network that uses convolutional layers to process sequential data. The convolutional layers extract significant features from the data. By using dilated convolutions, the TCN is able to capture long-term dependencies in a computationally efficient manner, making it a popular and efficient choice for temporal prediction tasks. 

Each of the above models and other classifiers will output a probability distribution as in \eqref{eq:Gamma} and a prediction according to \eqref{eq:Gamma_max}. However, the maximum probability for a specific object class may not be sufficiently high in order to have a high certainty prediction. Nevertheless, one may exploit the continuous time frame in which the hand manipulates a certain object and rapidly acquire additional samples while being certain that they originate from the same object. Hence, a majority vote can be used to accumulate probabilities over time \cite{Kahanowich2021}. Let $\ve{p}_t=\Gamma_\psi(S_{w.i}^{(t)})$ be the probability distribution over the $m$ object at time instance $t$. An accumulated prediction after $T$ time frames is the object $\mathcal{O}_{i^*}$ that acquires the maximum sum of scores given by
\begin{equation}
    \label{eq:majority}
    i^*=arg\max_i \sum_{t=1}^T \ve{p}_t.
\end{equation}
In this way, the model can keep accumulate and improve its prediction over time.

\section{Experiments}

\subsection{Experimental Setup}
An experimental system was constructed based on the two opposing fingers of the OpenHand Model-O \cite{Ma2017YaleOP} underactuated hand as seen in Figure \ref{fig:hand}. The experimental system consists of the hand and an automated reset mechanism. Once the object drops from the fingers, a thin string that runs through a hole in the center of the object pulls it into the reach of the fingers toward a new random grasp. A camera and a fiducial marker on the object are used solely to confirm the successful initial grasp in each episode. An action is defined to be $\ve{a}\in\{0^\circ,1^\circ,-1^\circ\}\times\{0^\circ,1^\circ,-1^\circ\}$ where, in each step, the actuators are stalled or moved by a constant amount. The system is operated by the Robot Operating System (ROS).

\begin{figure}
    \centering
    \includegraphics[width=\linewidth]{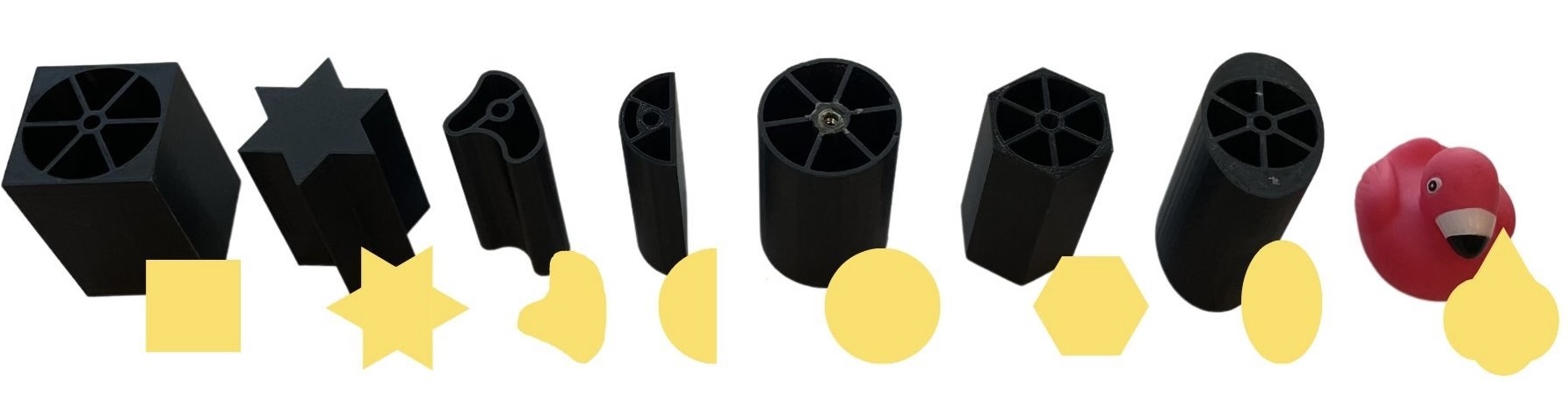}
    \caption{\small Eight distinct objects used for training the classification model.}
    \label{fig:test_objects}
\end{figure}

Eight objects, seen in Figure \ref{fig:test_objects}, are used for training and evaluating the classification model. Seven of the objects are prismatic PLA ones with different cross-sections including square, star-like, arbitrary-curved, half-circular, circular, hexagonal and elliptical. The eighth object is a rubber duck. Data was collected for the object where, in each episode, the hand grasps the object, performs an in-hand manipulation with a pre-defined action sequence for 10 seconds and then drops the object. The hand has no tactile sensors while only actuator angles and torques can be measured. During manipulation, data stream of these features is available in 10 Hz and recorded including instantaneous actions. 

\begin{table}[ht]
    \caption{\small Mean recognition success rate for various feature combinations and models }
    \label{tab:peak_accuracy}
    \centering
    \setlength{\tabcolsep}{6pt} 
    \begin{tabular}{lccccc}
        \toprule
        \multirow{2}{*}{Feature combination ($n$)}  & LSTM & TCN & FC-NN & RF & SVM\\
         & (\%) & (\%) & (\%) & (\%) & (\%) \\
        \midrule
        Torques (2) & 38.88 & 52.55 & 49.75 & 62.42 & 48.27\\
        Torques and actions (4) & 65.54 & 76.8 & 73.04 & 74.73 & 69.94\\
        Torques and angles (4) & 90.02 & 94.86 & 93.49 & 93.88 & 82.57\\
        Torq., ang. and act. (6) & 83.06 & 94.03 & 87.64 & 92.35 & 79.66\\
        \bottomrule
    \end{tabular}
    \label{tab:feature_combos}
\end{table}

\subsection{Object Recognition}

We begin by evaluating the classification accuracy over the eight objects. Dataset $\mathcal{D}$ was collected over $N_e=5,010$ episodes with approximately 625 episodes per object. The dataset was processed with the full length of the sequences, i.e., $w=100$, yielding a training dataset $\mathcal{P}$ with $N=5,010$ labeled sequences and 501,000 state samples. An additional test dataset of $860$ labeled sequences was collected and not included in the training set in any way.

We compare between several classification models including LSTM and TCN. While these are designated for temporal modeling, we also compare general classifiers including Random Forest (RF), Support Vector Machine (SVM) and a fully-connected NN (FC-NN). The hyper-parameters of the models were optimized yielding the following architectures. The TCN consists of three layers, two with 256 neurons, one with 128 neurons and a ReLU activation in between. A dropout of 10\% and an L1 regularizer with a factor of $10^{-3}$ were included to reduce over-fitting. Learning rate scheduling was also utilized, starting with a learning rate of 0.0001 and terminating at a rate of $10^{-6}$. The LSTM also consists of three layers consisting of 128, 64 and 32 neurons. A dropout of 5\% and L1 regularizer with a factor of $10^{-2}$ were used to decrease over-fitting. A learning rate scheduler was also used, initializing at 0.001 and terminating at a rate of $10^{-6}$. While LSTM and TCN can receive sequential input, the input for RF, SVM and FC-NN is a single vector. Hence, each sequence $S_{w,i}$ was flattened to a vector of size $wn$. The Random Forest consisted of 100 decision trees. The FC-NN consists of two layers of 128 and 64 neurons. A dropout of 10\% and L1 regularizer with a factor of $10^{-4}$ were used. A learning rate scheduler was also used, initialized at 0.001 and terminated at a rate of $10^{-6}$.

\begin{figure}
    \centering
    \includegraphics[width=\linewidth]{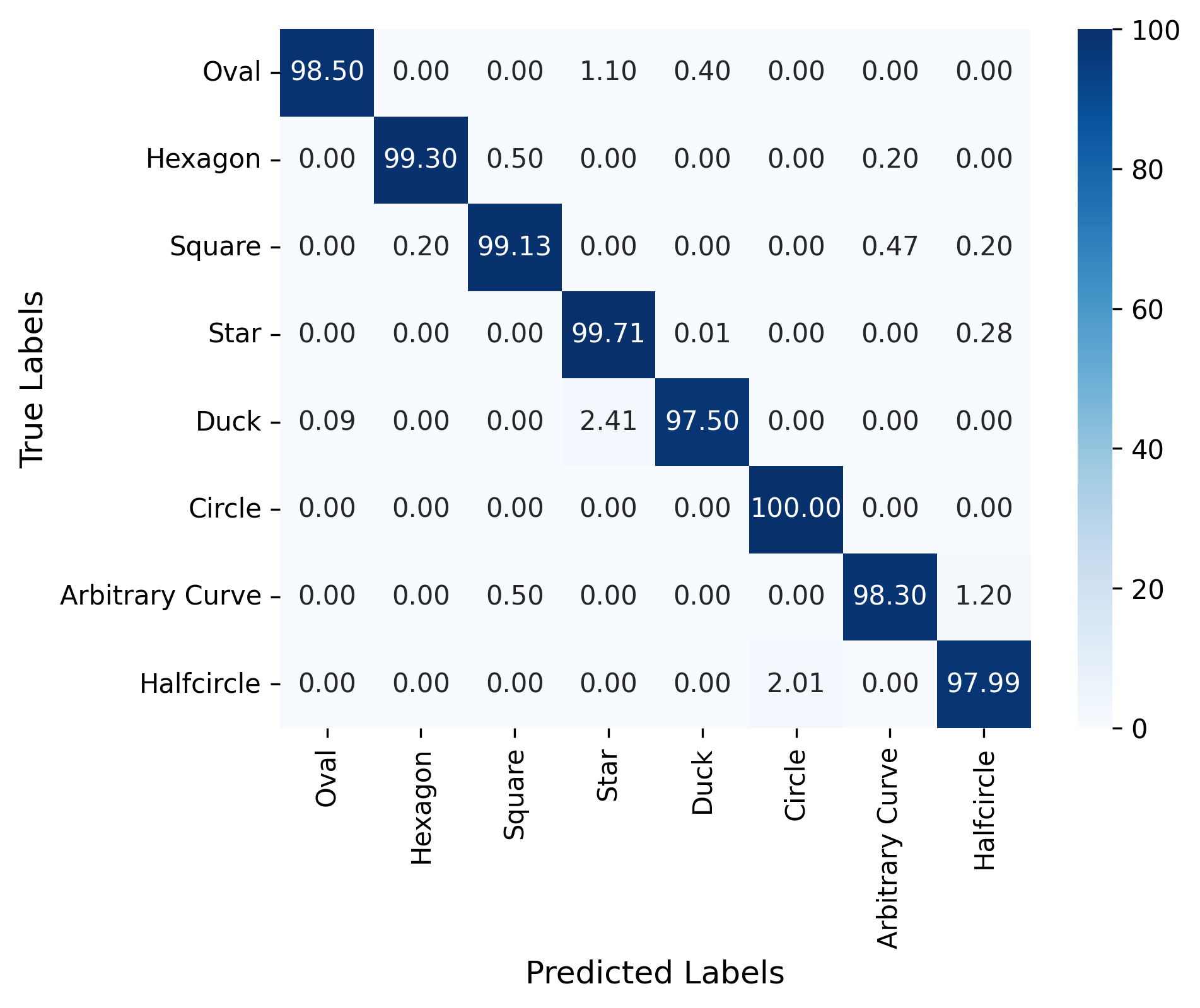}
    \caption{\small  Confusion matrix for classification of the eight objects with the TCN model over the test data.}
    \label{fig:tcnconfmat}
\end{figure}

Table \ref{tab:feature_combos} summarizes the results of the four models over the test data with four varying feature combinations of actuator angles, torques and actions. First, it is clear that torque signals along the motion are not sufficient in order to characterize the shape of the manipulated object. Including the actions provides marginal improvement since they are rather sparse. However, the best results are achieved with full state of the actuators having torques and angles. This four-dimensional state formulation was used in subsequent testing. When comparing between the five classification models, TCN with torques-angles state achieves the highest success with marginal advantage compared to FC-NN and RF. The results show that even simpler classifiers can successfully recognize object during in-hand manipulation. SVM provides inferior results to all models. Figure \ref{fig:tcnconfmat} exhibits the confusion matrix for classifying the eight objects with the TCN model. Results show high recognition rate for all objects in the set.

\begin{figure}
    \centering
    \includegraphics[width=\linewidth]{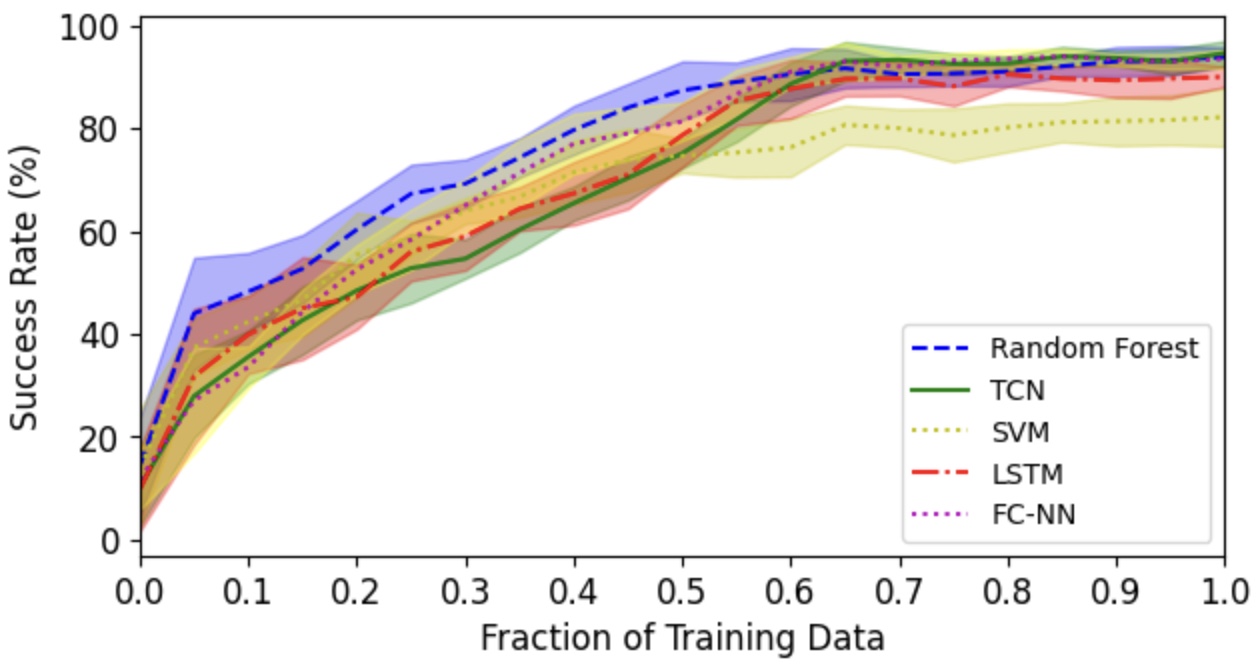}
    \caption{\small Object recognition success rate with regards to the fraction of total training data used to train RF, LSTM, SVM, FC-NN and TCN.}
    \label{fig:traindata}
\end{figure}

We next observe performance of the models with regards to the size of the training data. Recall that the full size of the training data is $N=5,010$ sequences amounting to approximately 16 hours of automated data collection. For a fair analysis, models were trained on varying portions of the data where, in each evaluation, data is picked from $\mathcal{D}$ without any shuffling so to maintain sequential order. For each data size, the model is re-trained 20 times on different parts of the data and the success rate is averaged. Results for success rate with regards to portion of the data in $\mathcal{P}$ used to train the model can be seen in Figure \ref{fig:traindata}. Success rate saturation is reached for all models with about 60\% of the dataset which corresponds to approximately 10 hours of data collection. TCN reaches the highest value in saturation whereas RF has higher values with a lower amount of training data. SVM performed the poorest and is not used in further evaluations. 

\begin{figure}
    \centering
    \includegraphics[width=\linewidth]{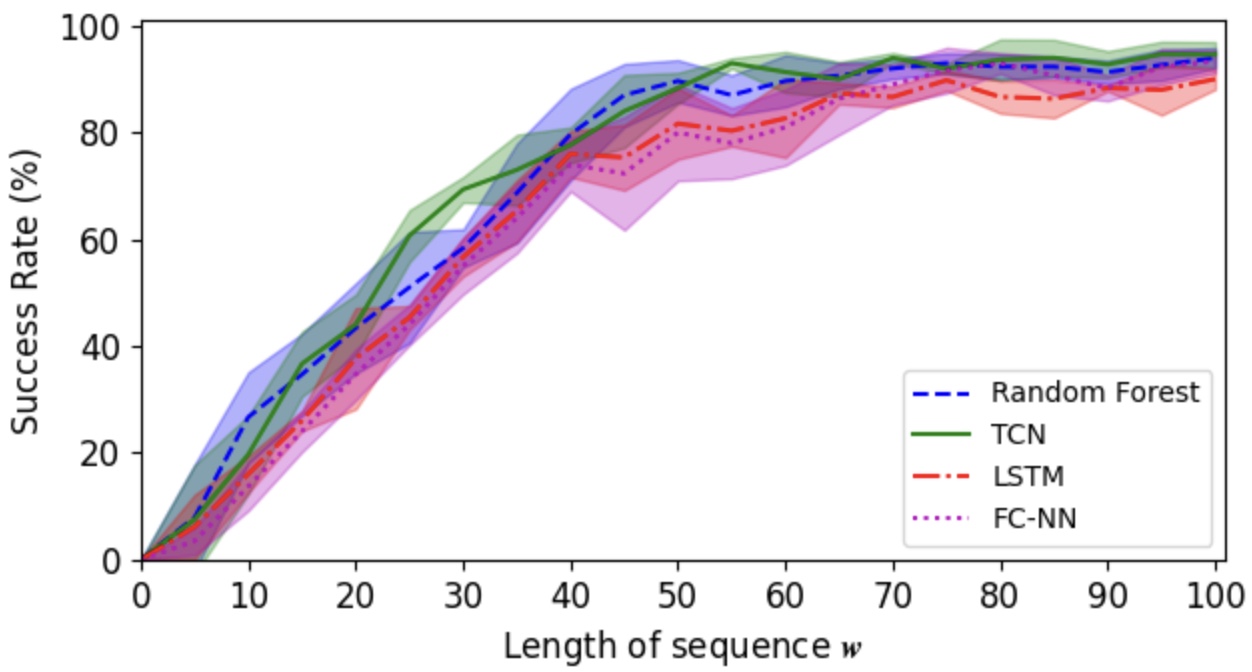}
    \caption{\small Object recognition success rate with regards to the length of the sequence $w$ in training and testing models TCN, LSTM, FC-NN and RF.}
    \label{fig:subsequences}
\end{figure}

The above trained models used sequence length $w=100$ for comparison. Analysis is now presented on the performance with regards to the value of $w$. Here also, for each value of $w$, the model was re-trained 20 times and the success rate was averaged. Figure \ref{fig:subsequences} depicts the success rate with regards to the length of the sequence $w$ over the test data. All models saturate with a sequence length $w>75  $. This result shows that sufficient motion must be exerted in order to acquire informative data for accurate recognition.

The use of the majority vote in order to accumulate certainties over objects and improve success rates is further evaluated. For the majority vote, we use a sequence size of $w=75$ and slide it over 25 inferences in order to accumulate probability distributions according to \eqref{eq:majority}. The majority vote is compared to single inferences with sequences of length $w=75$ and $w=100$. Table \ref{tb:majority} presents the results over four classification models. The results show marginal improvement when using the whole episode length of $w=100$ over a shorter one $w=75$ in a single inference classification. However, exploiting the entire episode but with a majority vote over sequences of $w=75$ provides extremely high success rates. Hence, even an ill-trained classifier can be used in a majority vote setting in order to improve predictions. In general, it is more advantageous to train a model with shorter sequence length and with majority vote, rather then having single inference with a longer sequence. The performance of the majority vote in real-time and with various sequence lengths will be analyzed later on.

\begin{table}
    \caption{\small Classification success rate of the eight objects for single inferences and with majority votes}
    \label{tb:majority}
    \centering
    \setlength{\tabcolsep}{5pt} 
    \begin{tabular}{llcccc}
        \toprule
         &  & LSTM & TCN & RF & FC-NN \\
         &  & (\%) & (\%) & (\%) & (\%) \\
        \midrule
        Single inference & $w=75$  & 89.80 & 93.17 & 91.67 & 93.00\\
        Single inference & $w=100$ & 90.02 & 94.86 & 93.88 & 93.49\\
        Majority vote & $w=75$     & 93.60 & 99.90 & 98.99 & 99.67\\
        \bottomrule
    \end{tabular}
\end{table}


\subsection{Generalized Recognition}

\begin{figure}
    \centering
    \includegraphics[width=\linewidth]{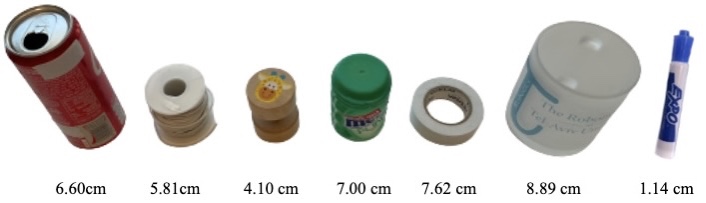}
    \caption{\small Circular household objects used for testing the generalization ability of the trained classifiers and their radii.}
    \label{fig:generalization_test_objects}
\end{figure}

The ability of a model to classify everyday objects into shape categories and
not only specific objects is now evaluated. We consider four shape categories: circular (7 objects), semi-circular (6 objects), square (9 objects) and elliptical (7 objects) prisms. A set of household test objects of varying sizes and within the categories was collected for benchmarking the generalization ability of the trained models. An example for the set of circular objects is seen in Figure \ref{fig:generalization_test_objects}. A successful recognition is the one that matches the shape category of a test object to the corresponding train object (seen in Figure \ref{fig:test_objects}). Table \ref{tb:generalization_accuracies} compares success rates between single inferences and majority vote for the shape categories. The classification of entire sequences ($w=100$) proved to be the least efficient in generalization, less so than classifying sequences of length $w=75$. The majority vote proves to be resilient in overcoming shape and size differences. Table \ref{tb:generalization_size_ranges} presents the size ranges with respect to the train objects that reached a success rate above 75\% and the actual success rates for majority votes of $w=75$ sequences. The results show a distinctive ability for the trained models to recognize shapes of new objects not included in the training and with different sizes.

\begin{table}
    \centering
    \caption{\small Classification success rate of four shape categories for single inferences and with majority votes}
    \label{tb:generalization_accuracies}
    \begin{tabular}{lccccc}
        \toprule
         &  & LSTM & TCN & RF & FC-NN \\
         &  & (\%) & (\%) & (\%) & (\%) \\
        \midrule
        Single inference & $w=75$  & 80.63 & 87.76 & 95.50 & 94.42\\
        Single inference & $w=100$ & 76.88 & 87.38 & 91.75 & 90.61\\
        Majority vote & $w=75$     & 89.59 & 94.55 & 98.00 & 96.04\\
        \bottomrule
    \end{tabular}
\end{table}
\begin{table}
    \centering
     \caption{\small Size range with respect to the training objects and success rate for recognizing the shape category of new objects with the TCN classifier }
    \label{tb:generalization_size_ranges}
    \begin{tabular}{lcc}
        \toprule
        Shape category & Size Range (\%) & Success rate (\%)\\
        \midrule
        Circular        & 32.2--193.7 & 91.5 \\
        Semi-Circular   & 56.0--181.5 & 95.5 \\
        Square          & 43.5--135.3 & 86.0 \\
        Elliptical      & 79.3--145.6 & 94.0 \\
        \bottomrule
    \end{tabular}
   
\end{table}
\begin{figure}
    \centering
    \includegraphics[width=\linewidth]{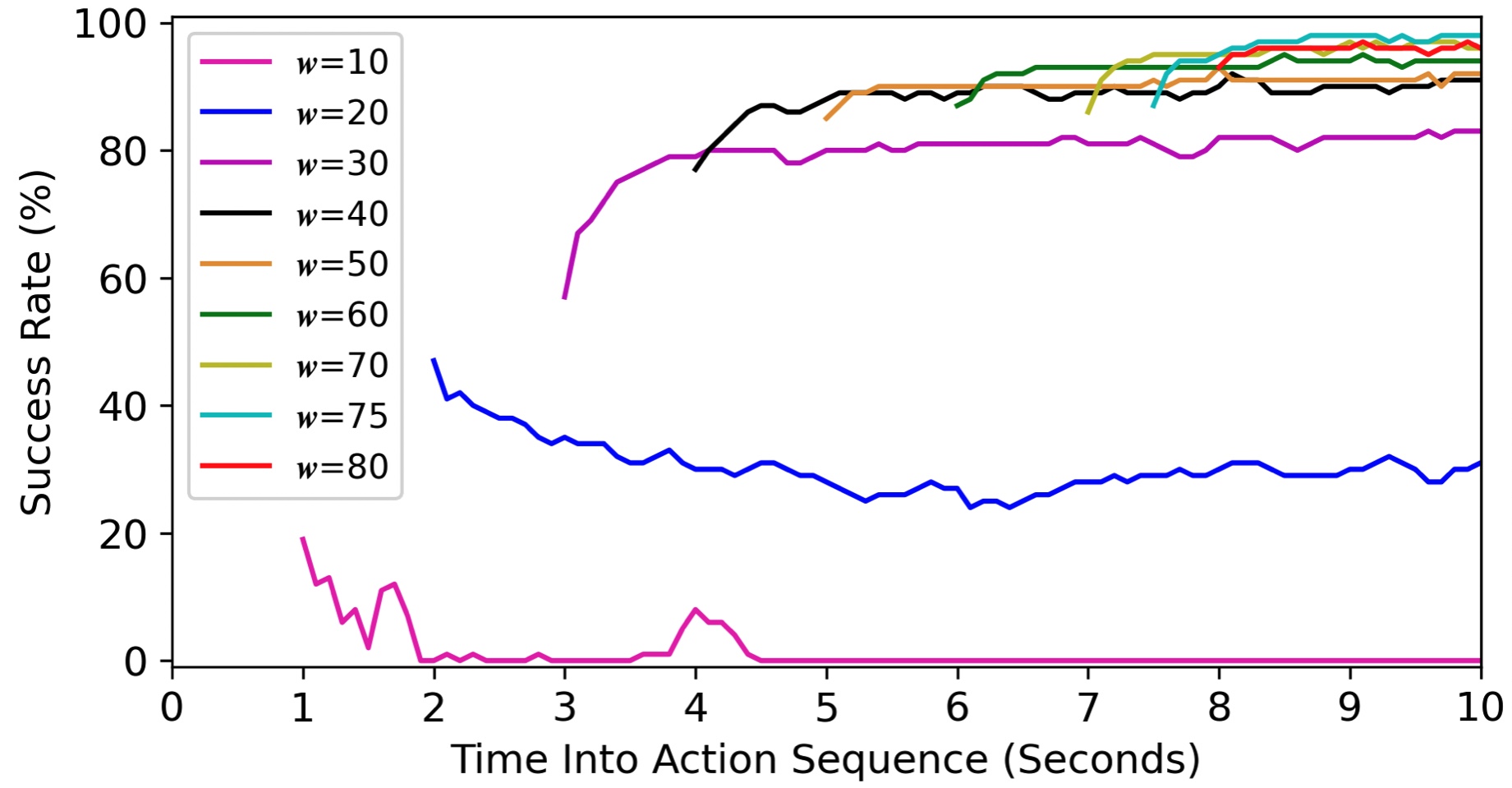}
    \caption{\small Success rate of shape recognition in during the manipulation of various objects using real-time majority votes.}
    \label{fig:real_time_accuracies}
\end{figure}
\begin{figure}[h]
    \centering
    \includegraphics[width=\linewidth]{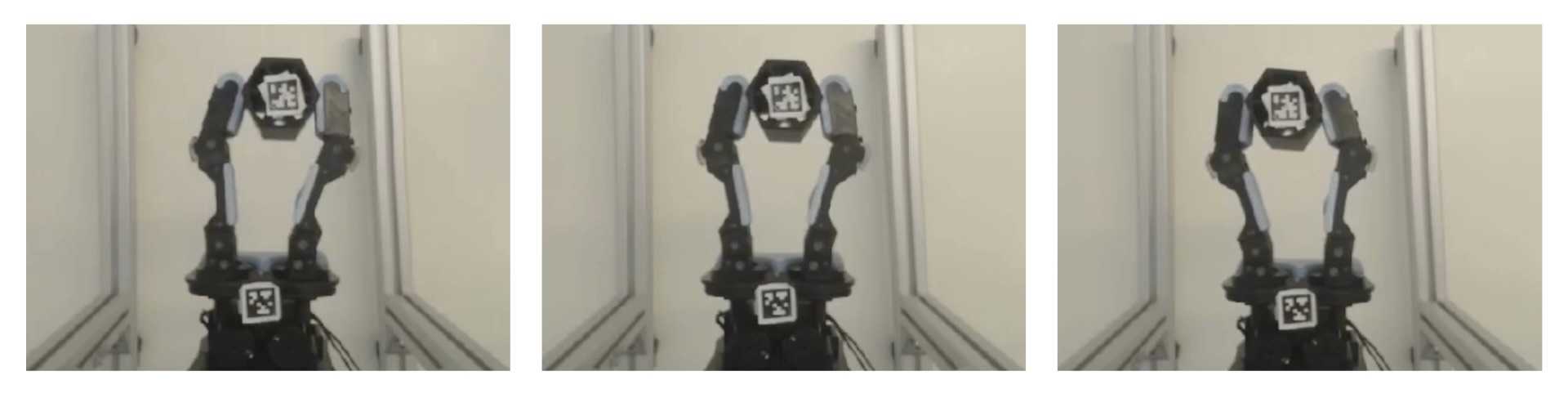}
    \caption{\small Snapshots of real-time object recognition of an hexagon prism over 25 time steps and 2.5 seconds. The model certainty of grasping the prism after 80, 90 and 100 time steps is 0.898, 0.914 and 0.920, respectively.}
    \label{fig:majority_vote_1}
\end{figure}
\begin{figure}[h]
    \centering
    \includegraphics[width=\linewidth]{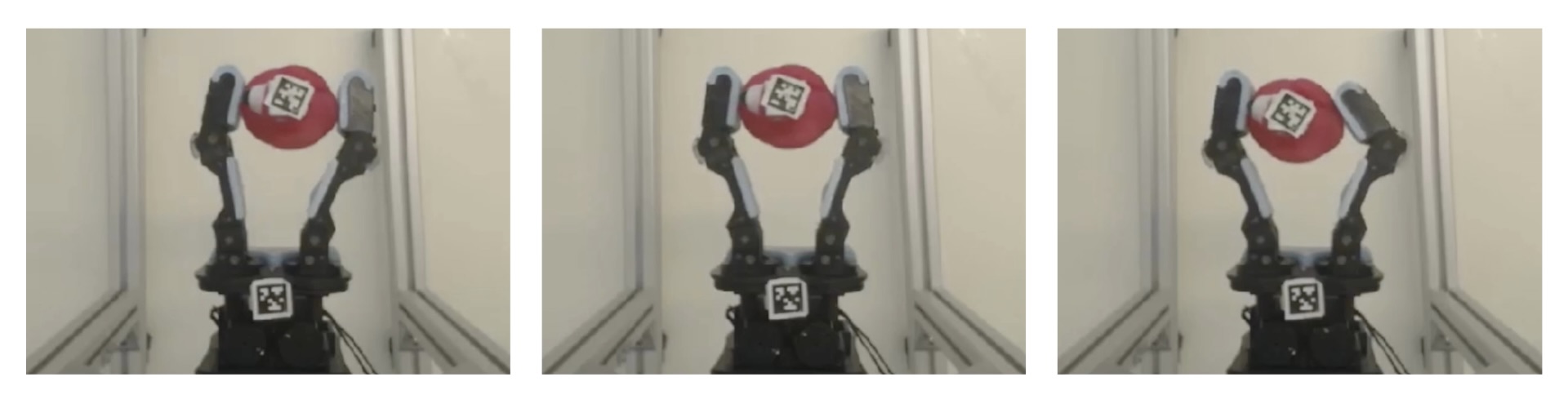}
    \caption{\small Snapshots of real-time object recognition of a rubber duck over 25 time steps and 2.5 seconds. The model certainty of grasping the duck after 80, 90 and 100 time steps is 0.923, 0.961 and 0.969, respectively.}
    \label{fig:majority_vote_2}
\end{figure}
\begin{figure}[h!]
    \centering
    \includegraphics[width=\linewidth]{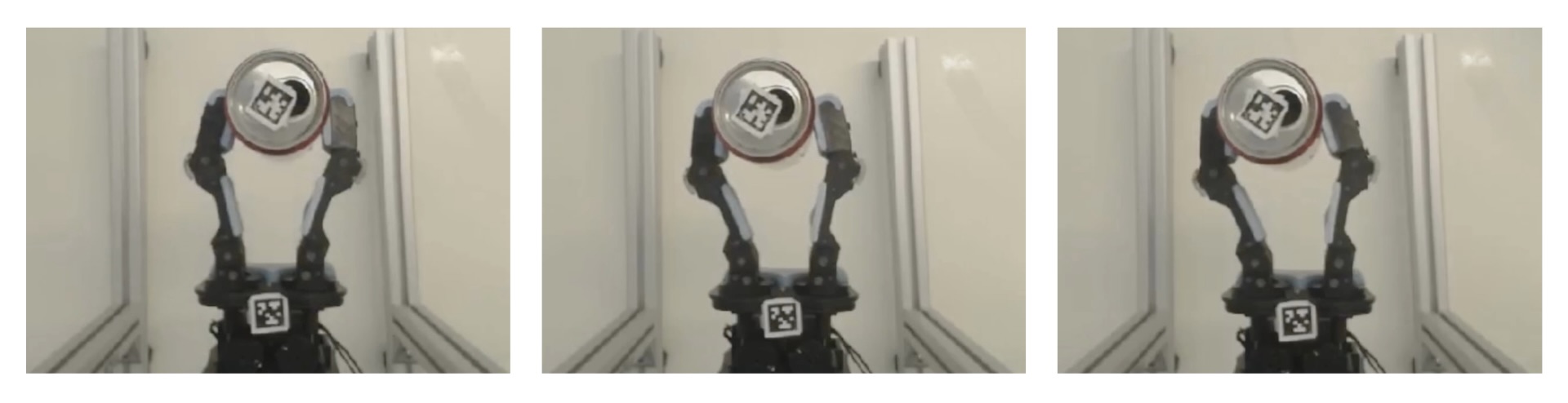}
    \caption{\small Snapshots of real-time object recognition of a soda can over 25 time steps and 2.5 seconds. The model certainty of grasping the can after 80, 90 and 100 time steps is 0.726, 0.776 and 0.812, respectively.}
    \label{fig:majority_vote_3}
\end{figure}


\subsection{Real-Time Object Recognition}

In the last experiment, we evaluate the ability of the model to conduct live object recognition early into the manipulation sequence and the prediction improvement with the majority vote. While we discuss real-time inference, we simulate real-time inference along the recorded test episodes and evaluate recognition success rate, focusing particularly on the model's ability to conduct generalized real-time recognition having been trained on a set of generic objects. Since a recorded episode is of length 10 seconds or 100 steps, a majority vote would have $T=100-w$ instances to collect predictions and make inferences according to \eqref{eq:majority}. Let $S_w^{(t)}$ be a sequence of length $w$ arriving at time $t$. Hence, once the first sequence $S_w^{(1)}$ arrives, an initial inference can be computed. Then, as additional states arrive, more sequences are acquired $\{S_w^{(2)}, S_w^{(3)},\ldots,S_w^{(T)}\}$ and predictions can be improved by updating the majority vote in \eqref{eq:majority}.

Figure \ref{fig:real_time_accuracies} presents the success rate analysis with the updating of the majority vote along time for a series of trials of general shape recognition. The analysis compares between different lengths of sequences used in inferences for the majority voting. First, having $w\leq20$ yields a decline in success rate. This is due to the significantly ill-trained and faulty classifier trained over short sequences with insufficient information. The conditions for a classifier to guarantee prediction improvement in a majority vote are given in \cite{sintov2023simple}. Classifiers with longer sequences are shown to improve predictions over time while reaching saturation at some point. Longer sequences, however, can provide a first prediction later in the episode. Sequences of approximately $w=75$ provide the highest success rate almost from the first inference with 88\% to saturation at 98\%. Figures \ref{fig:majority_vote_1}-\ref{fig:majority_vote_3} show snapshots of real-time object recognition inferences with three different objects. The motion that the object-hand system has to exert is rather small before a high certainty recognition is acquired. Hence, the hand can quickly move on to performing the desired task.

\section{Conclusions}

In this work, we have explored the ability of a tendon-based compliant hand to recognize grasped objects without any tactile sensing nor visual perception. We have shown that using only kinesthetic haptics during in-hand manipulation can provide sufficient information regarding the shape of the objects. In other words, with only position and torque data of the hand's internal actuators during motion with the objects, one can train a classifier to distinguish between the objects. A set of classification models were benchmarked including temporal ones. The results show best performance using the TCN with a marginal advantage over simple FC-NN or RF. In addition, majority vote with constant inferences during the hand-object motion is shown to increase prediction certainty and success rate. The proposed approach is also able to recognize a geometry characteristic of an object leading to a more general recognition model. The findings of this study have the potential to augment the capabilities of low-cost and 3D printed underactuated hands without any requirement for additional sensing hardware. 

\bibliographystyle{IEEEtran}
\bibliography{ref}

\end{document}